\documentclass[letterpaper, 10 pt, conference]{ieeeconf}  

\IEEEoverridecommandlockouts                              

\usepackage{graphics} 
\usepackage{graphicx}
\usepackage{epsfig} 
\usepackage{mathptmx} 
\usepackage{times} 
\usepackage{amsmath} 
\usepackage{amssymb}  
\usepackage{amsfonts}
\usepackage{bbold}
\usepackage{multirow}

\usepackage{layouts}
\usepackage{subfiles} 
\usepackage{xcolor}
\usepackage{booktabs}
\usepackage{xurl}
\newcommand{\ra}[1]{\renewcommand{\arraystretch}{#1}}
\usepackage{hyperref}
\usepackage{siunitx}
\usepackage{makecell}
\DeclareMathAlphabet{\mathcal}{OMS}{cmsy}{m}{n}
\SetMathAlphabet{\mathcal}{bold}{OMS}{cmsy}{b}{n}

\usepackage{tablefootnote}

\title{\LARGE \bf
Benchmarking Different QP Formulations and Solvers for Dynamic Quadrupedal Walking
}

\author{Franek Stark$^{1}$ and Jakob Middelberg$^{1,2}$ and Dennis Mronga$^{1}$ and Shubham Vyas$^{1,2}$ and Frank Kirchner$^{1,2}$
\thanks{This work was done in the AAPLE (grant number 50WK2275) and M-Rock (grant number 01IW21002) projects funded by the German Federal Ministry for Economic Affairs and Climate Action (BMWK) and the Ministry of Education and Research (BMBF) and is supported with funds from the federal state of Bremen for setting up the Underactuated Robotics Lab (201-342-04-2/2024-4-1)}
\thanks{Special thanks goes to Hannah Isermann and Rohit Kumar for their support in developing control software for quadrupedal walking.}
\thanks{$^{1}$All authors are with the Robotics Innovation Center at the German Research Center for Artificial Intelligence (DFKI), Bremen, Germany, Corresponding author's email: {\tt\small franek.stark@dfki.de}}%
\thanks{$^{2}$Jakob Middelberg, Shubham Vyas, and Frank Kirchner are additionally affiliated with the University of Bremen, Bremen, Germany}%
}

\DeclareMathSymbol{\shortminus}{\mathbin}{AMSa}{"39}
\usepackage[acronym]{glossaries}
\glsdisablehyper

\newacronym{mpc}{MPC}{Model Predictive Control}
\newacronym[plural=QP,firstplural=Quadratic Programs]{qp}{QP}{Quadratic Program}
\newacronym{srbd}{SRBD}{single rigid-body dynamics}
\newacronym{gs}{GS}{Gait Sequencer}
\newacronym{slc}{SLC}{Swing Leg Controller}
\newacronym{wbc}{WBC}{Whole-Body Control}
\newacronym{sfpw}{SFPW}{Solve Frequency per Watt}
\newacronym{cpu}{CPU}{Central Processing Unit}
\newacronym{hw}{HW}{hardware}
\newacronym{com}{COM}{Center of Mass}
\newacronym{ipm}{IPM}{Interior-point method}
\newacronym{asm}{ASM}{Active-set method}
\newacronym{admm}{ADMM}{Alternating direction method of multipliers}
\newacronym{alm}{ALM}{Augmented Lagrangian Method}
\newacronym{csc}{CSC}{Compressed Sparse Column}
\newacronym{pmm}{PMM}{Proximal Method of Multipliers}

\newacronym{fpga}{FPGA}{Field Programmable Gate Arraz}

\newacronym{hpipm}{HPIPM}{High Performance Interior Point Method}
\newacronym{osqp}{OSQP}{Operator Splitting Quadratic Programming Solver}
\newacronym{qpoases}{qpOASES}{Quadratic Programming Online Active Set Solver}
\newacronym{daqp}{DAQP}{Dual Active Set Solver for Quadratic Programming}
\glsunset{hpipm}
\glsunset{osqp}
\glsunset{daqp}
\glsunset{qpoases}

\begin{document}

\maketitle
\thispagestyle{empty}
\pagestyle{empty}

\begin{abstract}
\glspl{qp} are widely used in the control of walking robots, especially in \gls{mpc} and \gls{wbc}.
In both cases, the controller design requires the formulation of a \gls{qp} and the selection of a suitable \gls{qp} solver, both requiring considerable time and expertise. 
While computational performance benchmarks exist for \gls{qp} solvers, studies comparing optimal combinations of computational \gls{hw}, \gls{qp} formulation, and solver performance are lacking.
In this work, we compare dense and sparse \gls{qp} formulations, and multiple solving methods on different \gls{hw} architectures, focusing on their computational efficiency in dynamic walking of four-legged robots using \gls{mpc}.
We introduce the \gls{sfpw} as a performance measure to enable a cross-hardware comparison of the efficiency of \gls{qp} solvers. 
We also benchmark different \gls{qp} solvers for \gls{wbc} that we use for trajectory stabilization in quadrupedal walking. 
As a result, this paper provides recommendations for the selection of \gls{qp} formulations and solvers for different \gls{hw} architectures in walking robots and indicates which problems should be devoted the greater technical effort in this domain in future.
\end{abstract}

\glsresetall
\glsunset{hpipm}
\glsunset{osqp}
\glsunset{daqp}
\glsunset{qpoases}

\section{INTRODUCTION}



The development of quadrupedal robots has progressed rapidly in recent years, and several platforms have now reached a level of industrial maturity~\cite{hutter_anymal_2017,boston_dynamics_boston_2024,ghost_robotics_ghost_2024}. 
Apart from advancements in actuation, the main progress has been made on numerical methods for trajectory optimization, and its online implementation \gls{mpc}. 
At the ICRA 2022 conference, \qty{68}{\percent} of the papers presented at the legged robotics workshop covered receding horizon control or \acrshort{mpc}~\cite{katayama_model_2023}. 
Today, the de facto standard approach for legged locomotion comprises the planning of a contact sequence, computing a corresponding \gls{com} trajectory using \gls{mpc}, and, finally, stabilizing the obtained motion in real-time using \gls{wbc}~\cite{carpentier_recent_2021}. 

\begin{figure}[t]
    \centering
    \includegraphics[width=0.95\linewidth]{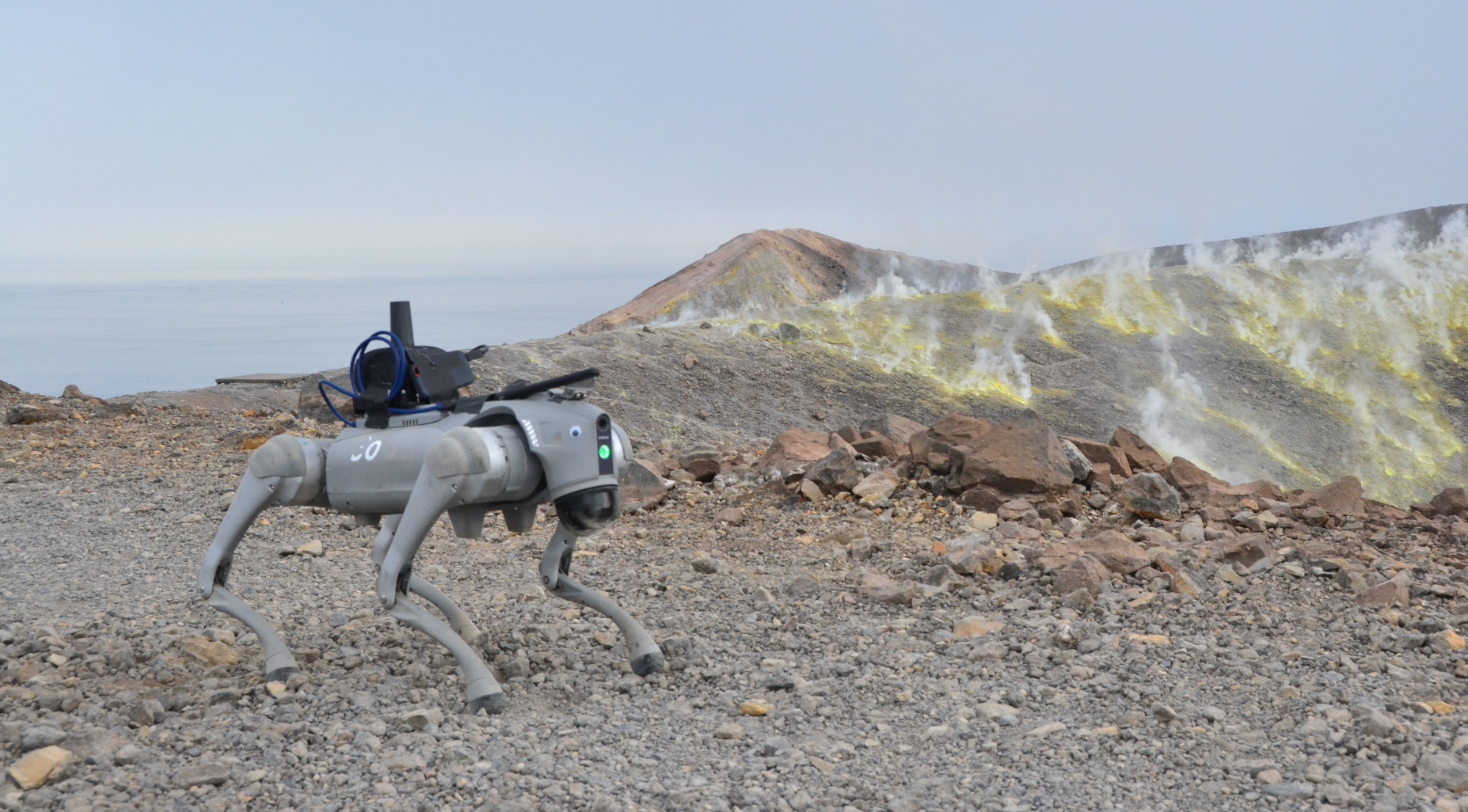}
    \caption{Go2 quadruped~\cite{unitree_robotics_unitree_nodate} used for evaluation in a Volcanic field test in an analogous scenario with limited onboard computing power.}
    \label{fig:go2_quad}
    \vspace{-0.5cm}
\end{figure}

At the core of \gls{mpc}, an optimal control problem has to be solved, which is formulated as linearly constrained \gls{qp}. 
Solving constrained \gls{qp}s is such a fundamental problem in both, \gls{mpc} and \gls{wbc}, that a huge amount of research has been devoted to developing efficient and stable \gls{qp} solvers, most of which are based on the \gls{asm}, \gls{ipm} or \gls{alm}. 
Each solver and each method is best suited for specific applications, e.g., some solvers show their strengths with small \gls{qp}s on embedded systems~\cite{ferreau_qpoases_2014}, others are optimized for larger problems that may arise in machine learning applications \cite{stellato_osqp_2020, boyd_distributed_2011}, and again others exploit the sparsity in the \gls{qp}~\cite{frison_hpipm_2020, stellato_osqp_2020}, a feature that can be advantageous in \gls{mpc} applications. 
Here, sparsity in the \gls{qp} is characteristic due to diagonal stacking of system state and control matrices over the entire prediction horizon. 
To reduce the \gls{qp} size, the state variables can be eliminated as decision variables, leading to denser and smaller matrices~\cite{jerez_sparse_2012}, a process that is known as \textit{condensing}. 
A great variety of problem formulations can be obtained by partially condensing the \gls{qp}, e.g., only eliminating half of the state variables~\cite{axehill_controlling_2015}. Special forms of condensed \gls{qp}s can be found in~\cite{jerez_sparse_2012} and~\cite{di_carlo_dynamic_2018}. 
In the works that introduce the current de facto standard of using \gls{mpc} and  
\gls{wbc}~\cite{di_carlo_dynamic_2018, kim_highly_2019}, the authors claim significant speed-up by removing some constraints and state variables from the \gls{qp}, leading to a dense and compact formulation. However, according to~\cite{axehill_controlling_2015}, sparse formulations are advantageous in \gls{mpc} as their computational and memory requirements grow linearly with the prediction horizon ($ \mathcal{O}(N) $) if sparsity is exploited, while for the dense variant, the computational cost is $\mathcal{O}(N^3)$ for \gls{ipm} and $\mathcal{O}(N^2)$ for \gls{asm}~\cite{dimitrov_sparse_2011}. Given these seemingly contradicting results, the effects of partial or full condensing of the \gls{mpc} problem with respect to the solver need to be investigated more thoroughly. 

There are several open-source benchmarks of \acrshort{qp} solvers~\cite{caron_qpbenchmark_2024, attila_kozma_benchmarking_2015}, which mostly use accuracy and solution time as a benchmark. However, as legged systems are being proposed for tasks such as space exploration~\cite{spiridonov_spacehopper_2024, arm_spacebok_2019}, other metrics such as power consumption and required onboard computing power become a critical factor for long-duration autonomous missions. ~\autoref{fig:go2_quad} illustrates the quadruped robot during a space exploration field test on a volcano, where it operates on limited battery capacity and onboard computing power to carry out the exploration tasks. Despite the importance of energy constraints, the impact of \gls{qp} formulation (sparse, partially condensed, condensed) and the relation to the prediction horizon, solver, and the computing \gls{hw} on performance used has yet to be comprehensively examined.

In this work, we focus on the application of dynamic quadrupedal walking using \gls{mpc}, which has so far produced several specialized methods for \gls{qp} formulation and solving, to reduce computation time or increase the planning horizon~\cite{ding_real-time_2019,di_carlo_dynamic_2018,kim_highly_2019}. 
We employ the standard approach to quadrupedal walking (contact planning, \gls{mpc}, \gls{wbc}) and evaluate it on a Unitree Go2 robot~\cite{unitree_robotics_unitree_nodate}. 
The benchmark involves (1) different \gls{qp} formulations in \gls{mpc} (sparse, partially condensed, and fully condensed), (2) two \gls{hw} architectures (x86, ARM) with desktop and single-board target computers, (3) various \gls{qp} solvers, including different principled methods for convex optimization, (4) different (dense) \gls{qp} solvers for \gls{wbc}. To allow a cross-\gls{hw} assessment of solver efficiency, we introduce the \textit{Solve Frequency per Watt} (SFPW) metric to compare different solvers. As a result of our benchmark, we recommend optimal combinations of computing \gls{hw}, problem sparsity, and \gls{qp} solvers for dynamic legged locomotion. The entire benchmark code, including the quadruped controller, is made open source\footnote{\url{https://github.com/dfki-ric-underactuated-lab/dfki-quad}}.

The remaining paper is structured as follows. Section~\ref{sec:dynamic_walking} describes the dynamic walking controller,  Section~\ref{sec:setup} the experimental setup and the performance metrics used for benchmarking. Section~\ref{sec:results} summarizes the results, with discussion in \autoref{sec:discussion}, and Section~\ref{sec:conclusions} draws conclusions on the benchmark.

\begin{figure}[t]
    \centering
    \includegraphics[width=0.99\linewidth]{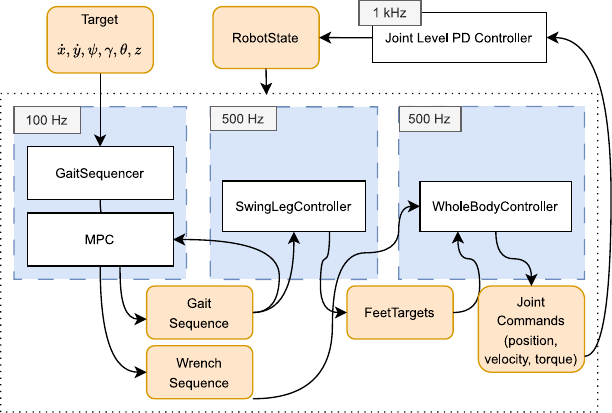}
        \caption{Block diagram showing dynamic walking controller's architecture}
    \label{fig:controller-architecture}
    \vspace{-1.5em}
\end{figure}

\section{DYNAMIC WALKING CONTROLLER}
\label{sec:dynamic_walking}

Our dynamic walking controller follows the general approach as described by \cite{di_carlo_dynamic_2018} and \cite{kim_highly_2019} and is illustrated in \autoref{fig:controller-architecture}. 
It outputs joint commands (position, velocity, torque) based on a target linear and rotational walking velocity and body posture.
The four sub-components of the controller run in parallel at different control rates: 
(1) The \textit{\gls{gs}} heuristically determines a gait sequence, containing the foot contact plan and the robot's target poses and velocities for the respective time steps over the prediction horizon $N$, 
based on the 
selected gait type and control target.
(2) The \textit{\gls{mpc}} calculates the optimal contact forces for the feet over the prediction horizon to reach the target pose and velocity specified in the gait sequence, assuming \gls{srbd}. 
(3) For the feet not intended for ground contact in the current time step, the \gls{slc} computes the trajectory using a Bézier curve.
(4) The \textit{\gls{wbc}} computes the optimal joint accelerations and torques, which achieve the given target positions and velocities for the feet, the body posture/velocity and contact forces, taking into account the whole-body dynamics.
\gls{gs} and \gls{mpc} run at \qty{100}{\Hz}, while \gls{slc} and \gls{wbc} run at \qty{500}{\Hz}.
The largest computing load is generated by \gls{mpc} and \gls{wbc}, which both require solving \glspl{qp} in real-time. These two components are detailed in the following.

\subsection{\acrlong{mpc}}\label{sec:method:mpc}
\subsubsection{Formulation}
The \gls{mpc} with $N$ prediction steps is formulated as the following convex optimization problem:
\begin{subequations}
\label{eq:mpc-form}
\begin{alignat}{7}
    \min_{\mathbb{X}, \mathbb{U}} && \sum_{k=1}^N  &\rlap{$(\mathbf{x}_{k} - \mathbf{x}_{k}^d)^T\mathbf{Q}~(\mathbf{x}_{k} - \mathbf{x}_{k}^d) + \mathbf{u}_{k-1}^T~\mathbf{R}~\mathbf{u}_{k-1}$} \\
    \text{s.t.~} && \mathbf{x}_{k+1} &= \mathbf{A} \mathbf{x}_{k} +\rlap{$\mathbf{B}_{k}\mathbf{u}_{k},$}&&&&k=0 \dots N-1 \label{eq:mpc-state-const}\\ 
    && \mathbf{f}^j_{m,k}&\leq \mathbf{f}_{k}^j \leq \mathbf{f}^j_{M,k},  &&&\forall j,~ &k=0\dots N-1\label{eq:mpc-u-bounds} \\
    &&|\mathbf{f}_{k}^j \cdot \hat{\mathbf{e}}_n| &\leq \mu\mathbf{f}_{k}^j \cdot \hat{\mathbf{e}}_z,&& \forall n\in\left\{x,y\right\},&\forall j,~ &k=0\dots N-1\label{eq:mpc-input-const}\\
    && \mathbf{x}_{0} &= \bar{\mathbf{x}}\label{eq:mpc-init_-const}
\end{alignat}
\end{subequations}
The decision variables $\mathbf{x}_{k} \in \mathbb{X},\mathbf{u}_{k} \in \mathbb{U}$, denote the prediction of the state and control input $k$ steps ahead at the current time step, $\mathbf{Q}$ is the diagonal state error cost matrix, $\mathbf{R}$ is the diagonal input cost matrix, and $\mathbf{x}^d_{k}$ the target state from \gls{gs}.
The equality constraint \eqref{eq:mpc-state-const} formulates the linearized system dynamics in world coordinates.
The state vector $\mathbf{x} \in \mathbb{R}^{13}$ contains the system's base orientation, position, linear, angular velocity, and the gravity constant. 
The control input $\mathbf{u} \in \mathbb{R}^{12}$ is a vector containing the four leg input forces $\mathbf{f}^j \in \mathbb{R}^3$, where $j\in \left\{1,2,3,4\right\}$ is the leg index.
The state matrix $\mathbf{A}$ is linearized around the current state $\bar{\mathbf{x}}$.
The input matrix $\mathbf{B}_{k} \in \mathbb{R}^{13\times12}$ consists of four $\mathbb{R}^{13\times3}$  blocks, mapping the respective feet contact force onto the system's state, depending on the planned contacts.
The bounding-box constraint \eqref{eq:mpc-u-bounds} limits the contact force to a maximum $\mathbf{f}_{\mathrm{M},k}^j$, enforces a minimum contact force $\mathbf{f}_{\mathrm{m},k}^j$ and sets it to zero if the respective leg $j$ is scheduled for swing phase at predicted step $k$.
The constraint \eqref{eq:mpc-input-const} keeps the contact forces within a linearized friction cone with coefficient $\mu$, where $\hat{\mathbf{e}}_x, \hat{\mathbf{e}}_y,$ and $\hat{\mathbf{e}}_z$ denote the standard unit vectors.
The equality constraint \eqref{eq:mpc-init_-const} sets the initial state to the current estimated state $\bar{\mathbf{x}}$.
\subsubsection{Partial Condensing}
Classically, \gls{mpc} problems are formulated into a \textit{dense} \gls{qp} by eliminating all state variables (except the initial state) as the state can be expressed as a function of the previous state and the input in LTI systems.
This work follows the idea of \cite{axehill_controlling_2015}, implemented by \cite{frison_efficient_2016}, where the number of eliminated state variables can be specified.
For this purpose, the original prediction horizon $N$ is divided into $N_p$ blocks, each containing several consecutive states.
The idea is that in each block, all states but the first one are eliminated by the system dynamics. 
Following this approach, a partially condensed \gls{qp} is formulated, which can be interpreted as another \gls{mpc} problem with a prediction horizon of $N_p$, enlarged input matrix, and input vector, but fewer states and dynamic constraints.
Further details can be found in \cite{frison_efficient_2016}.

\subsection{\acrlong{wbc}}\label{sec:method:wbc}
We use a variant of the \gls{wbc} in~\cite{del_prete_implementing_2016} to stabilize the quadrupedal walking. The \gls{wbc} considers the full system dynamics and solves for the joint accelerations $\ddot{\mathbf{q}} \in \mathbb{R}^{18}$ and contact forces $\mathbf{u} \in \mathbb{R}^{12}$ in a single QP: 
\begin{subequations}
 \label{eq:reduced_tsid}
\begin{alignat}{3}
\def\arraystretch{1.5}
 \underset{\ddot{\mathbf{q}}, \mathbf{u}}{\text{min}} && \quad \|\sum_i w_i(\mathbf{J}^i\ddot{\mathbf{q}} &+ \dot{\mathbf{J}}^i\dot{\mathbf{q}} - \dot{\mathbf{v}}^i_d)\|_2 + \|\mathbf{w}_f(\mathbf{u}_d-\mathbf{u})\|_2  \\
\text{s.t.} && \quad \mathbf{H}\ddot{\mathbf{q}} + \mathbf{h} &=  \mathbf{J}_{c}^T\mathbf{u}  \label{eq:reduced_tsid_1}\\
      && \quad \mathbf{J}_{c}\ddot{\mathbf{q}} &= -\dot{\mathbf{J}}_{c}\dot{\mathbf{q}} \label{eq:reduced_tsid_2}\\
      &&|\mathbf{f}^j \cdot \hat{\mathbf{e}}_n| &\leq \mu\mathbf{f}^j \cdot \hat{\mathbf{e}}_z, \forall n\in\left\{x,y\right\}, \mathbf{f}^j \cdot \hat{\mathbf{e}}_z > 0, \ \forall j  \label{eq:reduced_tsid_3}\\
       && \quad \boldsymbol{\tau}_{m} &\leq \mathbf{S}^{-1}\left(\mathbf{H}\ddot{\mathbf{q}} + \mathbf{h} - \mathbf{J}_{c}^T\mathbf{u}\right) \leq \boldsymbol{\tau}_{M}\label{eq:reduced_tsid_4}
\end{alignat}
\end{subequations}

where $w_i$ is the weight, $\mathbf{J}^i \in \mathbb{R}^{6 \times 18}$ the Jacobian, and $\dot{\mathbf{v}}_d^i \in \mathbb{R}^6$ the desired spatial acceleration for the $i$-th positioning task. The $\dot{\mathbf{v}}_d^i$ for the respective tasks are generated by PD-controllers to stabilize the body and feet trajectories produced by the \gls{mpc} and \gls{slc}. Further $\mathbf{u}_d \in \mathbb{R}^{12}$ the desired contact forces, as optimized by the \gls{mpc} in (\ref{eq:mpc-form}), i.e. $\mathbf{u}_d = \mathbf{u}_{k=0}$. The WBC can correct it to account for the full system dynamic. The amount of correction is specified by the diagonal weight matrix $\mathbf{w}_f$. 
The constraint (\ref{eq:reduced_tsid_1}) considers the rigid body dynamics, where $\mathbf{H} \in \mathbb{R}^{18 \times 18},\mathbf{h} \in \mathbb{R}^{18}, \mathbf{J}_{c} \in \mathbb{R}^{12 \times 18}$ are the mass-inertia matrix, bias vector, and contact Jacobian. 
Note that we consider only the floating base dynamics here 
, as we do not have torques as optimization variables in the \gls{qp}. 
The constraint (\ref{eq:reduced_tsid_2}) ensures that the feet' contacts are rigid and slip-free, while (\ref{eq:reduced_tsid_3}) prevents the contact forces from leaving the linearized friction cones, where $\mu$ is the friction coefficient. Finally, (\ref{eq:reduced_tsid_4}) ensures that the torque limits $[\boldsymbol{\tau}_m,\boldsymbol{\tau}_M]$ are respected, where $\mathbf{S}\in\mathbb{R}^{18\times12}$ is the actuator selection matrix. Note that in (\ref{eq:reduced_tsid_4}) we use the subscript $a$ to account only for the actuated joints. In contrast to~\cite{kim_highly_2019}, we use a weighting scheme for prioritization, as it is the common choice in WBC and is thus better for benchmarking. The resulting joint accelerations $\ddot{\mathbf{q}}$ are integrated twice and the joint torques are computed from $\mathbf{q},\dot{\mathbf{q}},\ddot{\mathbf{q}}$ by inverse dynamics. 

As alternative approach, we formulate the joint torques as additional optimization variables and solve for them directly in the QP. This leads to a larger QP, but avoids additional inverse dynamics computations.
We refer to this approach as \textit{full TSID} here, while (\ref{eq:reduced_tsid}) is called \textit{reduced TSID}. 

%

\section{EXPERIMENTAL SETUP}
\label{sec:setup}

\subsection{Performance Metrics}
\label{sec:performance}
Two different metrics are used to compare the performance of the \gls{qp} solvers:
(1) The solve time is the duration required for the solvers to solve the respective \gls{qp}.
(2) The \textit{\acrfull{sfpw}}, which we introduce to assess the efficiency of the respective \gls{qp} solvers independent of the computational \gls{hw}. This metric is inspired by the \textit{FLOPS per Watt} measure, employed by the Green500 list of the world's most power-efficient supercomputers \cite{hemmert_green_2010}:
\begin{gather}
    \mathrm{SFPW}=\frac{{\mathrm{solve~time}}^{-1}}{\mathrm{CPU~power~consumption}} \; \; \; \left[\frac{Hz}{W}\right]
    \label{eq:sfpw_calculation}
\end{gather} %

On x86, the Intel \texttt{RAPL} Interface determines the CPU power consumption.
For the ARM, the \texttt{tegrastats} utility is used. 
The CPU power consumption is sampled at \qty{10}{\hertz}.

The three target computers are listed in \autoref{tbl:computers}. 
The Jetson Orin NX and LattePanda Alpha are single-board computers suitable for installation on the quadruped due to their low power consumption and small form factor.

\begin{table}[t]
\centering
\caption{Target computers used for comparison.}
\label{tbl:computers}
\ra{1.3}
\begin{tabular}{@{}rllll@{}}\toprule
&\emph{Arch.} & \emph{CPU} & \emph{Cores}& \emph{RAM} \\
\midrule
Jetson Orin NX & ARM64 & AA78AEv8.2 & 8@\qty{2}{\giga\hertz}& \qty{16}{\giga\byte}  \\
LattePanda Alpha  & x86-64 & M3-8100Y & 2@\qty{3.4}{\giga\hertz} & \qty{8}{\giga\byte} \\
Desktop PC & x86-64 & i9-10900K & 10@\qty{3.7}{\giga\hertz} &\qty{16}{\giga\byte} \\
\bottomrule
\end{tabular}
\end{table}

\subsection{Implementation and Solvers}
We use the \textit{Unitree Go2} quadruped~\cite{unitree_robotics_unitree_nodate} for experimental evaluation. It is simulated using the Drake toolbox \cite{tedrake_drake_2019}.
The controller is implemented using ROS~2. 

The \gls{mpc} is implemented using the \gls{qp} interface of the acados framework \cite{verschueren_acadosmodular_2022}. 
It interfaces to a set of state-of-the-art QP solvers, including \gls{hpipm} \cite{frison_hpipm_2020}, which provides the (partial) condensing routines from \cite{frison_efficient_2016}.
The \gls{wbc} is implemented in the ARC-OPT framework~\cite{mronga_dennis_arc-opt_2024,mronga_whole-body_2022}, which also comes with a set of QP solvers and different WBC implementations, including the \textit{full} and \textit{reduced TSID} described in Section~\ref{sec:method:wbc}.
\begin{table}[htbp]\centering
\caption{ \gls{qp} Solvers compared in this work}
\label{tbl:solvers}
\ra{1.3}
\begin{tabular}{@{}rllcc@{}}\toprule
\emph{Solver} & \emph{Method} & \emph{Interface} & \emph{\acrshort{mpc}} & \emph{\acrshort{wbc}}\\
\midrule
 HPIPM \cite{frison_hpipm_2020} & \acrshort{ipm} & Sparse, Dense & \checkmark & \checkmark\\
 OSQP \cite{stellato_osqp_2020} & \acrshort{admm}$^*$ & Sparse & \checkmark & \\
 qpOASES \cite{ferreau_qpoases_2014} & parametric \acrshort{asm} & Dense & \checkmark & \checkmark\\
DAQP \cite{arnstrom_dual_2022} & dual \acrshort{asm} & Dense & \checkmark & \\
Eiquadproq \cite{buondonno_eiquadprog_2019} &  dual \acrshort{asm}$^\dagger$  & Dense & & \checkmark \\
PROXQP \cite{bambade_prox-qp_2022} & \gls{alm} & Dense & & \checkmark
 \\
\bottomrule
\multicolumn{5}{l}{\footnotesize{$^*$Variant of \gls{alm}, $^\dagger$Algorithm of Goldfarb and Idnani \cite{goldfarb_numerically_1983}}}
\end{tabular}
\end{table}
\autoref{tbl:solvers} lists all solvers that are considered in this work. 
A sparse interface means that a solver uses the sparse \gls{mpc} formulation as input and potentially exploits these.
In addition, the degree of sparsity of these solvers can be controlled via the partial condensation routines.
Some solvers only accept the fully dense \acrshort{mpc} formulation as input, which means that comparison in terms of sparsity is not possible.
We recognize that adjusting the hyperparameters, albeit time-consuming, can further improve performance. However, all solvers are used with the standard hyperparameters to ensure better comparability. The \acrshort{qpoases} solver is used with the \textit{\gls{mpc}} option set. The \gls{hpipm} solver has predefined modes that adapt the underlying \gls{ipm} algorithm. Here, the modes \textit{balanced} and \textit{speed\_abs} are used and treated as two different solvers. While the first mode provides more accurate results, the second focuses on speed \cite{frison_hpipm_2020}, which is more suitable for smaller systems and is equivalent to the HPMPC solver \cite{frison_high-performance_2014}.
All solvers are warm-started if possible. 

\subsection{Problem Sizes}
For comparison, we choose a short ($N=10$) and medium ($N=20$) \gls{mpc} prediction horizon.
The \gls{qp} dimension for both \gls{mpc} and \gls{wbc} are shown in \autoref{tbl:qp_sizes}.
For the \gls{mpc} solvers with sparse interfaces, this work compares all condensing levels such that the prediction horizon of the condensed \gls{qp}s are $N_p\in \{N,N-1,\dots,1\}$. 
The prediction horizon of the original \gls{mpc} is thereby evenly distributed into different blocks of size $\lfloor \frac{N}{N_p} \rfloor$. If $N$ is not an integer multiple of $N_p$, the remainder of $\frac{N}{N_p}$ is distributed to the foremost blocks (one per block). 
The final state is always left out during partial condensing and stays at size 1.
For the \gls{mpc} solvers with the dense interface, the \gls{qp} is fully condensed (including the terminal state), which is equivalent to partial condensing with $N_p=0$.
Note that the size of the constraint matrices in \gls{wbc} will change dynamically when changing the contacts during walking.
\begin{table}[htbp]\centering
\caption{\gls{qp} problem sizes}
\label{tbl:qp_sizes}
\ra{1.3}
\begin{tabular}{@{}rllccc@{}}\toprule
& & & \emph{Decision-} & \multicolumn{2}{c}{\emph{Constraints}}\\
&  & & \emph{variables} & \emph{Equality} & \emph{Inequality}\\
\midrule
\emph{\gls{wbc}} & \textit{Red. TSID}& & 30 &  18 & 28\\
\emph{\gls{wbc}} & \textit{Full TSID}& & 42 &  30 & 28\\
\midrule
\emph{\gls{mpc}} &\textit{original} & $N=10$ & 263 &  143 & 280\\
\emph{\gls{mpc}} &\textit{original} & $N=20$ & 513 & 273 &  360 \\
\multirow{2}{*}{\emph{\gls{mpc}}} &\multirow{2}{*}{\textit{condensed}} & & $13N_p + 12N $  & $13N_p + 13$ & $ 28N $  \\
& & &  $+~13$ &  &  \\
\bottomrule
\end{tabular}
\vspace{-1em}
\end{table}



\subsection{Test Scenarios}
Two distinct scenarios are evaluated for each test case. 
In the first scenario, the controller performs a trotting gait, executing various velocity profiles with a total duration of \qty{\approx 40}{\second}. 
This induces dynamic motions with speeds up to \qty{0.5}{\meter\per\second} and rotational velocities up to \qty{40}{\degree\per\second}. 
In the second scenario the quadruped remains in standing mode, while responding to commanded body roll, pitch, and yaw movements up to \qty{\pm 30}{\degree}, as well as height adjustments of \qty{\pm 10}{\centi\meter}. 
Here, the  total duration is \qty{\approx 20}{\second}. If the robot falls, the attempt is repeated twice before the test case is counted as a \textit{failure}. The dynamic simulation is run on a different computer in real time as to not affect the measurements.

    
\section{EXPERIMENTAL RESULTS}
\label{sec:results}
\subsection{\gls{mpc} Solve Time Analysis}
\begin{figure*}[htbp]
\centering
\includegraphics[width=0.99\linewidth]{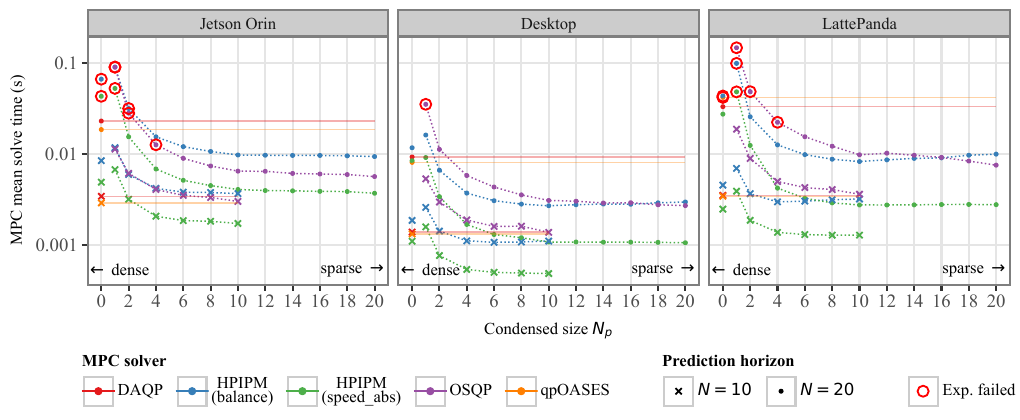}
\caption{Mean \gls{mpc} solution time for all solvers, target platforms and condensation levels for $N=20$ and $N=10$ in the trotting experiment.  
The qpOASES and DAQP solvers only show single points, as no comparison of condensation levels is possible here. Note that higher $N_p$ means the \gls{qp} is sparser.}
\label{fig:mpc_solve_time}
\end{figure*}
The mean \gls{mpc} solve time for the different solvers, target platforms and condensing levels for the \textit{trotting} experiment is depicted in \autoref{fig:mpc_solve_time}. The plotted time includes the condensing time, which is less than \qty{1}{\milli\second}.
Overall, solve time increases with increasing density of the \gls{qp}. 
For $N=20$, the dense formulation takes too long to solve for certain solvers leading to failed experiments, as indicated by red circles in \autoref{fig:mpc_solve_time}.
For the sparse solvers that support different levels of condensing (\gls{hpipm}, \acrshort{osqp}),  for $N_p \leq \frac{N}{2}$ the solve time increases exponentially with higher density, while it is approximately constant for $N_p > \frac{N}{2}$. 
The dense solvers (\acrshort{qpoases}, \acrshort{daqp}) perform better for fully condensed \gls{qp}s, but are outperformed by solvers that exploit sparsity.
As expected, all solvers show lower solve times for $N=10$ than for $N=20$. For all computer architectures, condensing levels, and horizons, the best-performing solver is \gls{hpipm} speed\_abs. 
For $N=20$, the second-fastest solver is \acrshort{osqp}, followed by \gls{hpipm} balance, although the difference is marginal on x86 systems.
\acrshort{daqp} and \acrshort{qpoases} are much slower for this prediction horizon.
For $N=10$, however, both \acrshort{qpoases} and \acrshort{daqp} perform almost as well as \acrshort{osqp} with $N_p=20$ (sparse problem).
The x86 desktop architecture is the fastest among the target systems compared, with \gls{hpipm} speed\_abs achieving sub-\unit{\milli\second} solve times. 
The Jetson and the LattePanda perform similarly, with the LattePanda being only slightly faster.
\begin{figure}
    \centering
    \includegraphics[width=0.99\linewidth]{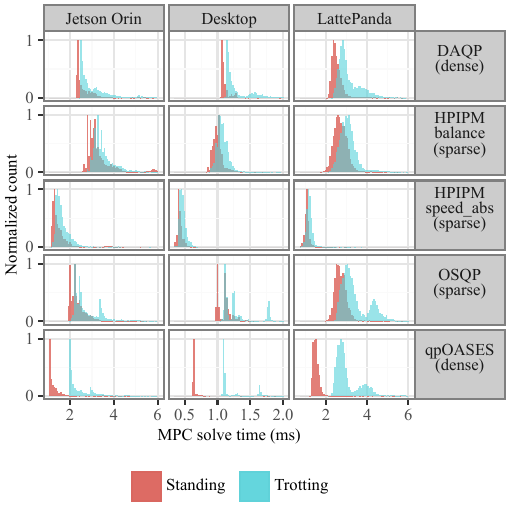}
    \caption{Histogram of the \gls{mpc} ($N=10$) solve times for selected solvers on all target platforms for dynamic trotting compared to standing. 
    }   \label{fig:mpc-solve-time-variance-n10}
\end{figure}
To compare the \gls{mpc} solve time between different test scenarios (standing, trotting), we use the normalized solve time histograms for all solvers on all computer architectures (see \autoref{fig:mpc-solve-time-variance-n10}). 
The histograms show that the mean solve time is less for the standing scenario than for trotting on all solvers. 
The most significant differences can be seen in \acrshort{daqp}, \acrshort{osqp} and \acrshort{qpoases}, where the histogram peaks are separated from each other. This effect is most prominent for \acrshort{qpoases}.
For the individual scenarios in themselves, the solution times for \gls{hpipm} are approximately normally distributed, while they show two significant peaks for the other solvers in trotting.
When comparing target computers, the slowest system (LattePanda) shows the highest variance in solve times.    


\subsection{WBC Solve Time Analysis}
\begin{figure}[!htpb]
    \centering
    \includegraphics[width=0.99\linewidth]{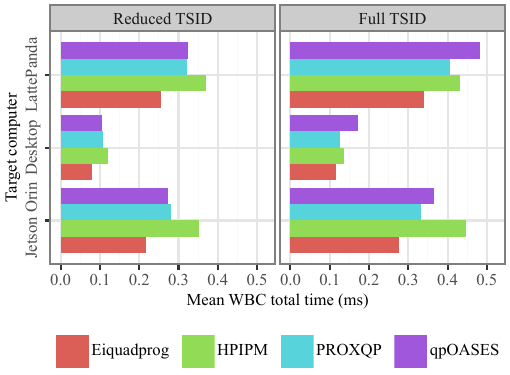}
    \caption{Computation time for different solvers using the \textit{reduced} and \textit{full TSID} \gls{wbc} in the trotting scenario.}
    \label{fig:wbc-total_time}
\end{figure}
\autoref{fig:wbc-total_time} shows the total computation time for one WBC cycle (set up time for the QP + solve time) using different solvers and two different WBC types. The set up time for the QP is solver-independent and thus not considered here separately.
The plots show that the Eiquadprog solver performs best in all comparisons. However, all solvers show a rather low computation time with $<0.5 ms$ on average. The computation time is slightly larger for the \textit{full TSID}, as compared to the \textit{reduced TSID} formulation on average.

\subsection{Efficiency Analysis}
\autoref{tbl:efficiency_all} shows the efficiency (measured in \gls{sfpw}) of different solvers in dynamic trotting over different target computers, planning horizons ($N=10$ and $N=20$), and \gls{wbc} formulations.
The most efficient solver per target computer is marked in bold.
The results the  show that the efficiency of the solvers per target system scales inversely with the solve time. 
Therefore, \gls{hpipm} speed\_abs has the highest efficiency on all systems for \gls{mpc} and Eiquadprog in case of the \gls{wbc}.

When comparing the efficiency of the three target computers, the Jetson Orin performs best: In case of \gls{mpc}, it is more than twice as efficient as the LattePanda and around three times as efficient as the desktop PC.
In case of \gls{wbc}, it is roughly four times faster than the others. 
\begin{table}[!htpb]
\centering
\caption{Efficiency of \gls{mpc} and \gls{wbc} solvers on different \gls{hw}}
\label{tbl:efficiency_all}
\ra{1.3}
\begin{tabular}{@{}cclccc@{}}\toprule
 \multicolumn{6}{c}{\emph{ Mean \gls{sfpw}} (\unit{\hertz\per\watt})} \\
\midrule
  & & & Jetson Orin & Desktop & LattePanda  \\ 
\cmidrule{2-6}
\parbox[t]{1mm}{\multirow{10}{*}{\rotatebox[origin=c]{90}{\gls{mpc}}}} & \parbox[t]{1mm}{\multirow{5}{*}{\rotatebox[origin=c]{90}{$N=10$}}} &\acrshort{daqp} & 77.47 & 17.41 & 24.69\\
&&\acrshort{hpipm} \scriptsize{balance} \scriptsize{(sparse)} & 70.08 & 22.01& 26.75 \\
&&\acrshort{hpipm} \scriptsize{speed\_abs} \scriptsize{(sparse)} & \textbf{159.40} & \textbf{50.80} &  \textbf{66.65} \\
&&\acrshort{osqp}  \scriptsize{(sparse)} & 90.35 & 17.68 & 24.22\\
&&\acrshort{qpoases} & 93.94 & 18.70 & 25.33\\
\cmidrule{2-6}
&\parbox[t]{1mm}{\multirow{5}{*}{\rotatebox[origin=c]{90}{$N=20$}}} &\acrshort{daqp} & \num{8.82} & \num{1.20} & \num{2.64}\\
&&\acrshort{hpipm} \scriptsize{balance} \scriptsize{(sparse)} & \num{22.35} & \num{7.50}& \num{8.80} \\
&&\acrshort{hpipm} \scriptsize{speed\_abs} \scriptsize{(sparse)} & \textbf{67.30} & \textbf{23.19} & \textbf{31.06} \\
&&\acrshort{osqp}  \scriptsize{(sparse)} & \num{42.79} & \num{8.58} & \num{11.77} \\
&&\acrshort{qpoases} & \num{11.36} & \num{2.46} & \num{2.20} \\
\midrule
\parbox[t]{1mm}{\multirow{8}{*}{\rotatebox[origin=c]{90}{\gls{wbc}}}} & \parbox[t]{1mm}{\multirow{4}{*}{\rotatebox[origin=c]{90}{Full TSID }}} & Eiquadprog & \textbf{958.93} & \textbf{212.66} & \textbf{257.58}\\
&&\acrshort{hpipm} & 576.14 & 177.92 & 200.99\\
&&Proxqp & 795.34 & 190.80 & 216.11\\
&&\acrshort{qpoases} & 713.63 & 142.82 & 180.80\\
\cmidrule{2-6}
&\parbox[t]{1mm}{\multirow{4}{*}{\rotatebox[origin=c]{90}{Red. TSID}}}  &Eiquadprog & \textbf{1259.37} & \textbf{304.99} & \textbf{319.45}\\
&& \acrshort{hpipm} & \num{753.73} & \num{198.35} & \num{234.55}\\
&& Proxqp & 954.99 & 231.86 & 265.41\\
&& \acrshort{qpoases} & 977.83 & 238.09 & 263.70\\
 \bottomrule
\end{tabular}
\end{table}


\section{DISCUSSION}
\label{sec:discussion}
The experimental results provide valuable insights into the performance of different solvers in dynamic quadrupedal walking regarding (1) sparsity, (2) variance over different tasks, and (3) \gls{hw} efficiency.

\paragraph{Sparsity} For larger \gls{qp}s, as they occur in \gls{mpc}, sparse formulations, when tackled by sparse solvers, generally seem to be advantageous over dense formulations. 
This effect becomes more significant with increasing prediction horizon. 
Here, condensing the \gls{qp} and solving it has the opposite effect: the solution time increases significantly. 
This finding is in line with the theory of \cite{frison_efficient_2016} and \cite{axehill_controlling_2015} that in \gls{mpc}, if the input vector has almost the same size as the state vector, the fully sparse formulation is a good choice.
For robots with fewer control inputs, however, condensing could improve performance.
The results also indicate that for even smaller prediction horizons than $N=10$ or smaller formulations, as in \cite{di_carlo_dynamic_2018}, \gls{asm} solvers such as \acrshort{qpoases} could be advantageous. 
The \gls{mpc} problem analyzed here seems to be at the limit of the problem size where \gls{asm} is outperformed by other methods. 
It should also be considered that the \gls{hpipm} speed\_abs solver, which outperforms the other solvers even at $N=10$, does not provide accurate results for some applications. 
In the case of \gls{wbc}, where significantly smaller problems are considered, the dense solvers and especially \gls{asm} perform well for the reduced formulation. 
The results for the full TSID formulation show that if the problem gets bigger, other methods such as \gls{ipm} could indeed be considered as an option.
In contrast to \gls{mpc}, the choice of the \gls{qp} solver is not of great relevance for \gls{wbc} problems. Instead, efforts should be directed towards researching robust and stable \gls{qp} formulations for \gls{wbc}.

\paragraph{Variance over different tasks} The comparison of different motion tasks shows that in static cases such as standing, the \gls{asm}-based solvers, especially \acrshort{qpoases}, has a lower solve time than when trotting. 
This is in line with the general finding that \gls{asm} benefit from a stable problem structure, as the active-set can be warm-started~\cite{kuindersma_efficiently_2014, bartlett_active_2000}. 
The same finding applies to \acrshort{osqp}.
\gls{ipm} solvers, here \gls{hpipm}, show a certain robustness against changing problem structures, as already stated by \cite{frison_hpipm_2020}. 
In contrast to all other solvers, \gls{hpipm} does not exhibit a second peak in the solve time histogram while the robot is trotting.
The compromise could be that, for robot tasks where the problem is very dynamic, \gls{ipm} provides constant performance, while for in static tasks \gls{asm} or \gls{alm} might be advantageous.

\paragraph{ \gls{hw} efficiency} The comparison of the different computer architectures shows that the solution times of the Jetson Orin and the Latte Panda are comparable, while the desktop delivers significantly faster solutions. 
However, the efficiency measurement shows that the faster solution times come at the expense of power consumption, resulting in a comparable efficiency for both x86 systems, with the LattePanda being more efficient in the range of $\qty{10}{\hertz\per\watt}$ depending on the prediction horizon and selected solver.
On the other hand, the Jetson Orin is at least twice as efficient as the Desktop, and for the fastest solver and $N=10$ even up to three times as efficient.
This makes the Jetson Orin and ARM an ideal platform for applications with energy constraints, which is almost always the case in legged robotics.
\section{CONCLUSIONS}
\label{sec:conclusions} 

This work describes a benchmark for \gls{qp} solvers in \gls{mpc} and \gls{wbc} in dynamic quadrupedal walking. 
The benchmark involves different \gls{qp} formulations (sparse to dense), several solvers, robot tasks and \gls{hw} architectures.

From this work's findings, three main conclusions follow: (1) For \gls{mpc}, sparse solvers and especially solvers based on \gls{ipm} (here \gls{hpipm}) perform best in dynamic quadrupedal walking and should be considered especially for long prediction horizons. These solvers also show certain robustness to changing problem structures, e.g., when changing contacts or between different tasks, and are therefore better suited for dynamic quadrupedal walking than other methods. 
(2) In \gls{wbc}, any of the regarded open-source solvers performs well; the engineering effort should rather be put into the formulation of the \gls{wbc} problem itself. 
(3) ARM architecture (here Jetson Orin) shows better efficiency than x86 when considering the \acrlong{sfpw} as a metric. Thus, they should be preferred in resource-constrained applications like autonomous quadrupedal walking.


Future work includes extending the benchmark to other \gls{hw} architectures (e.g. CUDA), additional solvers to verify the results obtained, and possibly more complex systems (e.g. humanoids) to investigate the impact of system complexity on performance.

        







\bibliographystyle{IEEEtran}
\bibliography{referenceszot}

\end{document}